\def\paperID{10857}
\def\confName{ICCV}
\def\confYear{2025}

\def\paperTitle{Uncertainty-Aware Diffusion-Guided Refinement of 3D Scenes}

\newif\ifreview 
\newif\ifarxiv 
\newif\ifcamera \newcommand{\cameraready}{\cameratrue}
\newif\ifrebuttal 

\cameraready 

\pdfoutput=1
\documentclass[10pt,twocolumn,letterpaper]{article}
\ifreview \usepackage[review]{iccv} \fi

\ifarxiv \usepackage[pagenumbers]{iccv} \fi
\ifrebuttal \usepackage[rebuttal]{iccv} \fi
\ifcamera \usepackage{iccv} \fi

\usepackage{graphicx}	
\usepackage{amsmath}	
\usepackage{amssymb}	
\usepackage{booktabs}
\usepackage{times}
\usepackage{microtype}
\usepackage{epsfig}
\usepackage{caption}
\usepackage{float}
\usepackage{pifont}
\usepackage{placeins}
\usepackage{color, colortbl}
\usepackage{stfloats}
\usepackage{enumitem}
\usepackage{tabularx}
\usepackage{xstring}
\usepackage{multirow}
\usepackage{xspace}
\usepackage{url}
\usepackage{subcaption}
\usepackage{xcolor}
\usepackage[hang,flushmargin]{footmisc}

\ifcamera \usepackage[accsupp]{axessibility} \fi

% ...

\newcommand{\name}[1][]{\texttt{UAR-Scenes}}

\newcommand{\xmark}{\ding{55}} 
\newcommand{\cmark}{\ding{51}}

\ifarxiv  \fi
\definecolor{deepgreen}{RGB}{0,128,0}

\newcommand{\R}[1]{{%
    \textbf{%
        \ifstrequal{#1}{1}{\textcolor{red}{R#1}}{%
        \ifstrequal{#1}{2}{\textcolor{blue}{R#1}}{%
        \ifstrequal{#1}{3}{\textcolor{magenta}{R#1}}{%
        \ifstrequal{#1}{4}{\textcolor{teal}{R#1}}{%
                           \textcolor{cyan}{R#1}%
        }}}}%
    }%
}}

\usepackage{xr-hyper}

\makeatletter
\newcommand*{\addFileDependency}[1]{
  \typeout{(#1)}
  \@addtofilelist{#1}
  \IfFileExists{#1}{}{\typeout{No file #1.}}
}

\makeatother

\definecolor{cvprblue}{rgb}{0.21,0.49,0.74}
\usepackage[pagebackref,breaklinks,colorlinks,allcolors=cvprblue]{hyperref}
\usepackage[capitalize]{cleveref}
\crefname{section}{Sec.}{Secs.}
\crefname{table}{Table}{Tables}
\crefname{figure}{Fig.}{Figs.}

\ifarxiv \crefname{appendix}{App.}{Apps.}
\else \crefname{appendix}{Suppl.}{Suppls.} \fi

\author{
Sarosij Bose, Arindam Dutta, Sayak Nag, Junge Zhang, Jiachen Li, \\Konstantinos Karydis, Amit K. Roy-Chowdhury\\
University of California, Riverside, USA\\
{\small \texttt{\{sbose007,adutt020,snag005,jzhan744,jiachen.li\}@ucr.edu, \{kkarydis,amitrc\}@ece.ucr.edu}}
}

\begin{document}
\title{\paperTitle}

\twocolumn[{
  \maketitle
  \centering
  \vspace{-1em}
  \begin{minipage}{\textwidth}
    \centering
    \includegraphics[width=1.01\linewidth,keepaspectratio]{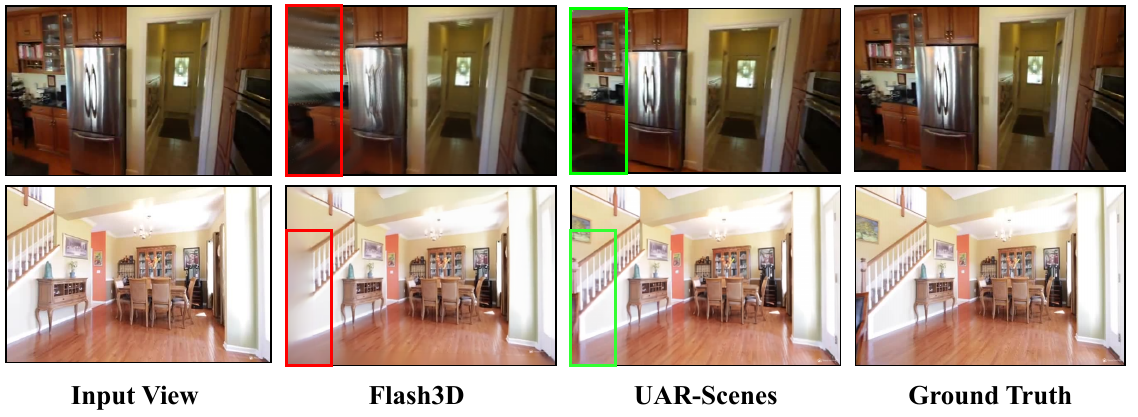}
  \end{minipage}
  \vspace{0.25em}
  \captionof{figure}{
    \textbf{Overview:} We introduce \name{}, 
    a diffusion-guided refinement pipeline that enhances outputs from pre-trained single-image to 3D scene reconstruction models, such as Flash3D~\cite{szymanowicz2024flash3d}, which yield imperfect renderings {\color[HTML]{FF0000}{(red box)}} under slight viewpoint variations (2nd column from left). By harnessing the generative power of a latent video diffusion model (LVDM)~\cite{blattmann2023stable}, our approach can sample plausible explanations for unseen regions {\color[HTML]{009901}{(green box)}} to produce refined, high-quality novel views of 3D scenes (3rd column from left).
  }
  \label{fig:teaser}
  \vspace{2em}
}]
\begin{abstract}

Reconstructing 3D scenes from a single image is a fundamentally ill-posed task due to the severely under-constrained nature of the problem. Consequently, when the scene is rendered from novel camera views, existing single image to 3D reconstruction methods render incoherent and blurry views. This problem is exacerbated when the unseen regions are far away from the input camera. In this work, we address these inherent limitations in existing single image-to-3D scene feedforward networks. To alleviate the poor performance due to insufficient information beyond the input image's view, we leverage a strong generative prior, in the form of a pre-trained latent video diffusion model, for iterative refinement of a coarse scene represented by optimizable Gaussian parameters. To ensure that the style and texture of the generated images align with that of the input image, we incorporate on-the-fly Fourier-style transfer between the generated images and the input image. Additionally, we design a semantic uncertainty quantification module that calculates the per-pixel entropy and yields uncertainty maps used to guide the refinement process from the most confident pixels while discarding the remaining highly uncertain ones. We conduct extensive experiments on real-world scene datasets, including in-domain RealEstate-10K and out-of-domain KITTI-v2, showing that our approach can provide more realistic and high-fidelity novel view synthesis results compared to existing state-of-the-art methods. The project page is available at: \url{https://github.com/UCR-Vision-and-Learning-Group/UAR-Scenes}
\end{abstract}
\section{Introduction}
\label{sec:intro}

3D Scene Reconstruction is a long standing challenging task, playing a crucial role in holistic scene understanding~\cite{zhang2021holistic, tang2024diffuscene},
generative content creation, mixed reality~\cite{yu2024wonderjourney}, robot navigation and path planning~\cite{matsuki2024gaussian, Jiang2023FisherRF, liang2022bevfusion,Matsuki:Murai:etal:CVPR2024}. 
Reconstructing 3D scenes involves Novel View Synthesis (NVS), wherein the scene is rendered from unseen camera angles to obtain novel views of the scene. 
Although notable progress has been made towards improving NVS, most existing methods rely on multiple images from numerous viewpoints~\cite{yu2021pixelnerf,charatan2024pixelsplat,chen2024mvsplat} to increase the available information, which may not always be available. 
This leads us to the relatively underexplored task of 3D reconstruction from a single RGB image~\cite{szymanowicz2024flash3d}. Although the existing single image to 3D scene method can generate reasonable results, it often fails when the scene is rendered from an unseen camera angle, due to scarcity of information limiting its capability to reconstruct scenes in views it has never seen before. This bottleneck leads us to ask the question: ``\emph{Can we design a pipeline that refines scenes so that it retains high fidelity in regions it has seen and generates consistent plausible completions for the unseen regions?"}

One way to compensate for this lack of multi-view information is to use generative priors so that they can provide valuable complementary information. The principle is to synthesize plausible 3D details even for regions not visible in the image, where \emph{plausibility} refers to the fact that visible portions should remain multi-view consistent, while occluded or out-of-view areas should integrate seamlessly with the context of the scene (refer to supplementary for qualitative results).
Some existing approaches toward solving this problem include text-guided inpainting~\cite{mirzaei2023reference} and distillation of 2D diffusion priors~\cite{wu2024reconfusion}. 
However, these methods either lack 3D correspondence awareness~\cite{wang2024innerf360} or do not have any temporal continuity as they denoise each image separately. This leads to issues with multi‐view consistency for large viewpoint changes as the camera moves further away from the source. One way to address this challenge is to 
reformulate the problem in a 2.5D reconstruction~\cite{wimbauer2023behind, li2024know} setting, which can be limited in terms of fully capturing unseen regions. 
Moreover, extending such approaches to fully unbounded real world scenes, where the scene may not loop back to the starting point, is non-trivial. These methods often rely heavily on just generated features, leading to mode collapse~\cite{poole2023dreamfusion}, content distortion~\cite{poole2023dreamfusion, qian2024magic123} and visible artifacts in rendered views~\cite{tang2024dreamgaussian, yi2024gaussiandreamer}, especially when the camera motion spans a large range. 
Although diffusion priors are well-suited for object‐level enhancement, these limitations indicate that priors primarily designed for object reconstruction, cannot be directly applied for scene refinement.

To this end, we propose \textit{\underline{U}ncertainty-\underline{A}ware Diffusion-Guided \underline{R}efinement of 3D \underline{Scenes}: \name{}}, %
an uncertainty‐aware Gaussian splatting based refinement framework that combines the strengths of regressive and generative approaches. 
We perform the refinement using \textit{Adaptive Densification and Pruning (ADP)}~\cite{matsuki2024gaussian}, a mechanism that controls the density of 3D points by introducing or deleting Gaussian primitives (check supplementary Section 6.1). ADP plays a crucial role in expanding or shrinking Gaussians depending on whether the region is over-constructed or not. Therefore, in observed regions where the confidence per pixel is high, our method mostly preserves Gaussian primitives with high‐fidelity details obtained from the reconstruction backbone, while in unobserved or occluded areas, it learns to introduce new Gaussians by using uncertainty-driven sampling from a Latent Video Diffusion Model (LVDM)~\cite{wang2024motionctrl} to produce plausible scene completions. 
We leverage an MLLM~\cite{li2023blip}, along with an open-vocabulary segmentation model~\cite{li2022languagedriven}, to compute the per-pixel entropy present in LVDM generated views which gives us the per-pixel uncertainty associated with the generated images. 
In this way, the inherent ambiguity present in single-view reconstruction methods can be addressed, expanding the applicability of our pipeline from simple object‐centric datasets or scenes with limited backgrounds~\cite{barron2022mip} to 
 arbitrarily large real‐world unbounded scenes~\cite{zhou2018stereo}. Furthermore, it does not rely on any form of ground truth supervision (2D or 3D), making the entire process self‐supervised and thus, post-hoc compatible with any existing 3D reconstruction algorithm. This design effectively addresses ambiguities, such as missing artifacts and inconsistent 3D reconstructions inherent in single‐view settings. Thus, in a nutshell, our proposed model  \emph{integrates a differentiable feed-forward scene reconstruction pipeline that operates from a single image with an uncertainty-aware LVDM for self-supervised refinement of noisy scenes}. 

The \textbf{main contributions} of our work are as follows: 
\begin{itemize}
    \item We propose \name{}, a novel uncertainty-aware refinement pipeline which can take an existing single image to 3D scene reconstruction model and generate reliable and plausible explanations for unseen and naturally occluded regions present in unbounded real-world scenes. To the best of our knowledge this is the first optimization-based work for the refinement of single image to 3D reconstruction methods. 

    \item \name{} utilizes a camera pose controlled Latent Video Diffusion Model (LVDM) which generates consistent posed images to serve as pseudo 2D supervision
    for refining a noisy scene. To address noisy pseudo-views, we design an innovative uncertainty quantification module  which leverages an MLLM guided open vocabulary segmentation model for highlighting the most confident pixels for \name{} to learn from. 
    \item To address any texture and style inconsistencies between the input image and the generated pseudo views from the LVDM, we propose a Fourier Style Transfer (FST)-based texture alignment methodology. 
    
    \item We conduct extensive experiments on both real-world in-domain dataset RealEstate-10K~\cite{zhou2018stereo} (on which the baseline method was trained) and an out-domain dataset KITTI-v2~\cite{geiger2012we} which is unseen for both our method and the baseline, demonstrating that \name{} enhances performance significantly.
\end{itemize}

\section{Related Works}
\label{sec:related}
\textbf{Regressive 3D Reconstruction from Sparse Views.} 
Existing generalizable reconstruction methods~\cite{szymanowicz2024flash3d, xu2024murf, chen2024mvsplat, szymanowicz2024splatter, wewer2024latentsplat, tang2025hisplat, xu2025depthsplat} aim to train a single feed-forward model which can predict a 3D scene directly. 
Such methods include PixelSplat~\cite{charatan2024pixelsplat} which utilizes epipolar transformers to reduce scale ambiguity and obtain better scene features. 
Similarly, MVSplat~\cite{chen2024mvsplat} builds cost volumes to obtain Gaussians in a fast and efficient manner. LatentSplat~\cite{wewer2024latentsplat} utilizes a GAN decoder as a generative component but GANs are known to be highly unstable~\cite{kodali2017convergencestabilitygans} and hence unsuitable for modeling real-world unbounded scenes. Recently, Flash3D~\cite{szymanowicz2024flash3d} introduced a single image to 3D scene pipeline leveraging a pre-trained depth estimator. In this paper, we build upon this regressive framework and show how most works falling in this paradigm fail to generate realistic completions for unseen regions. 

\noindent\textbf{Generative Scene Reconstruction and Synthesis.} 
Our method falls within an under-explored area of utilizing only 2D (pseudo) labels to supervise 3D tasks using latent diffusion models. Existing approaches like~\cite{anciukevivcius2023renderdiffusion, anciukevivcius2024denoising, tewari2023diffusionforwardmodelssolving, szymanowicz2023viewset} denoise multi-view images using a 3D-aware denoiser. However, such methods do not produce a coherent 3D structure since the denoising process is done separately for each image leading to sub-optimal multi-view consistency. Works such as 3DGS-Enhancer~\cite{liu20253dgs} address a similar problem in a sparse view setting but they employ a mix of refined views along with un-refined views when training the video-diffusion model.  Other works such as CAT3D~\cite{gao2024cat3d}, GeNVS~\cite{chan2023generative}, ReconFusion~\cite{wu2024reconfusion}, LM-Gaussian~\cite{yu2024lm}, and ReconX~\cite{liu2024reconx}\footnote{Code and models are not \underline{publicly available} for these methods. ReconFusion was trained on an internal dataset.} all tackle similar problems but they deal with the problem by employing expensive conditioning techniques~\cite{rombach2022high} which may not always be multi-view consistent. \textit{Further, none of these methods optimize the Gaussians directly but instead fine-tune an existing generative diffusion model on datasets to make it generalizable, which ties the models down to the particular distribution on which it is trained.}

\noindent\textbf{Uncertainty Quantification in 3D Reconstruction.} Quantifying uncertainty has received relatively less attention in 3D scene reconstruction, especially considering it's importance in the context of single image to 3D reconstruction where there is inherently a lot of ambiguity. One of the first approaches in estimating epistemic uncertainty in NeRFs from sparse views using variational inference was studied in works~\cite{shen2021stochasticneuralradiancefields,shen2022conditionalflownerfaccurate3d, savant2024modelinguncertaintygaussiansplatting} but owing to increasing parameters alongwith using different models, these approaches are inherently expensive. BayesRays~\cite{goli2023} proposed a method to quantify uncertainty in any NeRF based reconstruction module in a post-hoc fashion. Further, uncertainty has also been explored in the context of active-view selection and path planning in the field of 3D mapping in robotics~\cite{pan2022activenerf, Jiang2023FisherRF} to estimate which next view maximizes the information gain over a given pre-trained model by approximating the hessian matrix of the log-likelihood function. However, this becomes un-scalable to compute for each Gaussian primitive. Inspired by these approaches, we opt for a semantic scene understanding based approach for quantifying the uncertainty in novel test-time views for refinement. 
\begin{figure*}
    \centering
    \includegraphics[width=\linewidth]{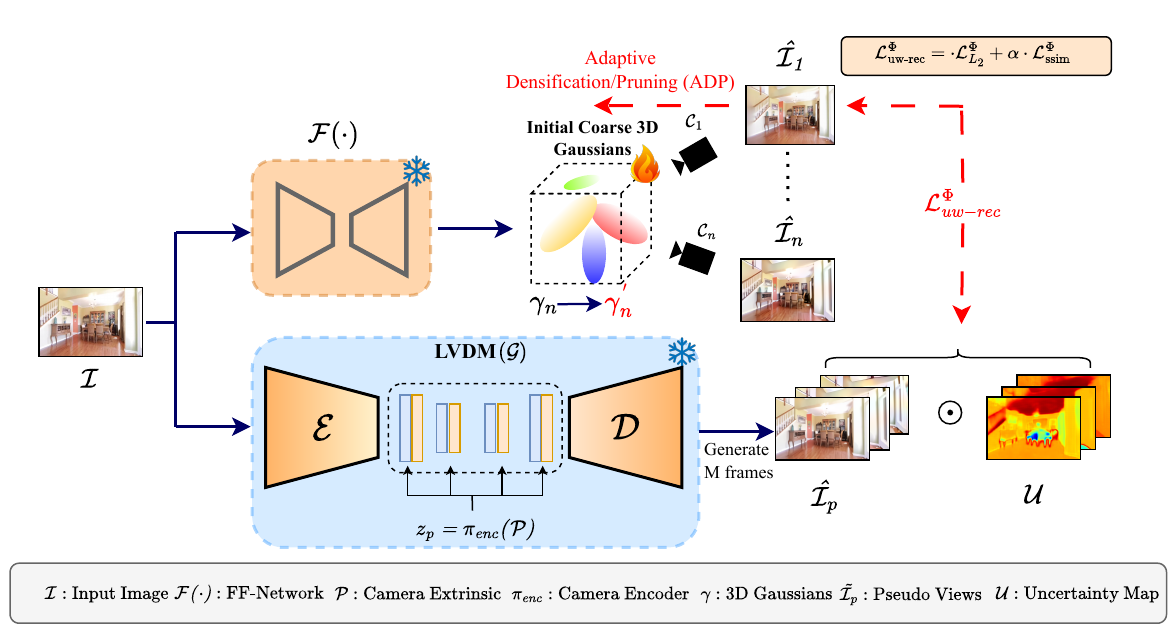}
    \caption{\textbf{Workflow of \name{}.} For a conditioning image $\mathcal{I}$, a pre-trained 3D reconstruction model $\mathcal{F(\cdot)}$ produces coarse gaussians ($\gamma_n$) representing the scene $\phi$, represented by optimizable gaussian parameters. Using temporally consistent pseudo 2D supervisory images ($\widetilde{\mathcal{I}}_{\text{p}}$) sampled from the pre-trained camera extrinsic embedded LVDM model~\cite{wang2024motionctrl}, we iteratively refine the gaussians $\gamma_n$ using \emph{Adaptive Densification and Pruning (ADP)}~\cite{kerbl20233d} to obtain clean gaussians ($\gamma^{'}_n$). To gauge the uncertainty of each pixel $\textbf{p}$ in the generated pseudo images ($\widetilde{\mathcal{I}}_{\text{p}}$), we propose a semantic uncertainty quantification method. We estimate the entropy present in each ($\widetilde{\mathcal{I}}_{\text{p}}$) obtained by utilizing an off-the-shelf open-vocabulary segmentation model $\mathcal{S}$~\cite{li2022languagedriven} using which we obtain uncertainty maps $\mathcal{U}$ (as shown in {\hyperref[subsec:coarse_init]{Figure~\ref{fig:uq}}}). We take the hadamard product between $\widetilde{\mathcal{I}}_{\text{p}}$ and $\mathcal{U}$ forming the target objective for the refinement loss in {\hyperref[eq:uw_fst__recon]{Equation~\ref{eq:uw_fst__recon}}}, which guides the refinement process.}
    \vspace{-1em}
    \label{fig:pipeline}
\end{figure*}
\section{Methodology}
\label{sec:method}

\textbf{Problem Formulation.} Consider a single input image $\mathcal{I}$ with camera pose intrinsic $\mathcal{K}$, a sequence of test-camera trajectory extrinsics $\mathcal{P}_{1\ldots M}$, and any given pre-trained deterministic feed-forward reconstruction model $\mathcal{F}(\cdot)$, from which we obtain $N$ noisy Gaussian splats each representing a scene. Our goal is to refine these $N$ noisy 3D Gaussian splats, $\phi_n \in \mathbb{R}^{N \times d}$ to better representations $\phi_n^{'} \in \mathbb{R}^{N \times d}$ leveraging the strong generative prior $\mathcal{G}$. 
We further sample $k$ supplementary views derived from the LVDM for providing pseudo 2D supervision, which refines the noisy Gaussians \emph{without access to any 2D or 3D ground truth}. 

\noindent \textbf{Overview.} We provide an overview of our uncertainty-sensitive refinement pipeline in {\hyperref[fig:pipeline]{Figure~\ref{fig:pipeline}}}. To get an accurate target representation of the scene, the geometric features from the backbone deterministic reconstruction model are preserved and the textures from the LVDM are utilized. 
The discussion starts with the deterministic model, serving as the initial point for refining scene Gaussians, detailed in {\hyperref[subsec:coarse_init]{Section~\ref{subsec:coarse_init}}}. Then, we briefly discuss the topic of camera controllable LVDMs in {\hyperref[subsec:lvdm_cam]{Section~\ref{subsec:lvdm_cam}}} since it is crucial to align the pseudo views and the noisy rendered views so that they are in the same camera space. The next part addresses the method of weighting the optimization objective based on the confidence associated with each pixel, as outlined in {\hyperref[subsec:uq]{Section~\ref{subsec:uq}}}. The final section describes the application of the video diffusion model to generate pseudo-views and facilitate iterative refinement, further elaborated in {\hyperref[subsec:refinement]{Section~\ref{subsec:refinement}}}.

\subsection{Initial Coarse 3D Representation}
\label{subsec:coarse_init}
Here, we briefly discuss the working formulation behind the 3D reconstruction model $\mathcal{F}(\cdot)$ which provides us with an initial set of gaussians representing the scene $\phi$. Flash3D~\cite{szymanowicz2024flash3d} uses an off-the-shelf monocular depth estimation network~\cite{piccinelli2024unidepth} to estimate the metric depth $\textbf{D}$ of the input image $\mathcal{I}$. It utilizes a U-Net~\cite{ronneberger2015u} with ResNet blocks~\cite{he2016deep} to produce multiple sets of depth-ordered layered Gaussians per pixel $\textbf{p}$ estimating to a certain degree the occluded regions of the scene. The model is then trained over a collection of such scenes and, for each scene, it renders images along a known set of camera extrinsics to get the corresponding rendered images $\mathcal{\tilde{I}}$. These paired rendered and ground truth views $(\mathcal{\tilde{I}}, \mathcal{I}_{\text{gt}})$ can then be used to supervise the initialization model with the reconstruction objective:

\begin{equation}
\mathcal{L}_{\mathrm{rec}} 
= \|\mathcal{\tilde{I}} - \mathcal{I}_{\text{gt}}\|_{2} 
+ \alpha\,\mathrm{SSIM}\bigl(\mathcal{\tilde{I}}, \mathcal{I}_{\text{gt}}\bigr),
\label{eq:f3d_recon}
\end{equation}

\noindent where $\mathcal{\tilde{I}}$ is the image rendered by the initialized model $\mathcal{F(\cdot)}$ rasterizing its layered Gaussians and $\mathcal{I}_{\text{gt}}$ is the corresponding ground-truth image at train time. $\alpha$ is a hyper-parameter weighting the SSIM~\cite{wang2004image} loss.

\noindent\textbf{Challenges.} As illustrated in {\hyperref[fig:teaser]{Figure~\ref{fig:teaser}}}, it is clear that relying solely on depth-based modeling to capture the structure and appearance of scenes is often insufficient in many cases. This issue becomes more pronounced in scenarios involving substantial camera motion, where a significant portion of the rendered image extends beyond the field of view of the input image. Moreover, merely utilizing the MSE loss as described in {\hyperref[eq:f3d_recon]{Equation~\ref{eq:f3d_recon}}} drives the model to average out pixels in unseen regions, which often results in suboptimal, blurry renderings.

\subsection{LVDM with Camera Control} 
\label{subsec:lvdm_cam}
Existing LVDMs (refer to the supplementary for the general formulation of LVDMs) cannot generate denoised images with user-specified camera pose trajectories. MotionCtrl~\cite{wang2024motionctrl} addresses this issue which we have used in our method. It injects the Pl\"{u}cker embedding information of the corresponding relative camera extrinsics: $\{\,x^T,\; x_{(R,T)},\; \textbf{R},\; \textbf{T}\}$ and embeds them into the cross-attention layers of the U-Net which results in a modified revised objective (refer to {\hyperref[eq:motionctrl_obj]{Equation~\ref{eq:motionctrl_obj}}} of the supplementary for original objective of LVDMs):

\begin{equation}
\resizebox{0.9\columnwidth}{!}{$
\mathbb{E}_{x_{0}^{1..M},\, y_{t}^{1..M},\, \epsilon_{t}^{1..M} \sim \mathcal{N}(\mathbf{0},I)}
\Bigl\|
\epsilon_{t}^{1..M} 
- \epsilon_{\theta}\bigl(z_{t}^{1..M}, t, y, \mathcal{C}_{(R,T)}\bigr)
\Bigr\|_2,
$}
\label{eq:motionctrl_obj}
\end{equation}

\noindent where $\mathcal{C}_{(R,T)}$ is the embedding of the camera extrinsics. Upon completion of the training process, the conditioning image $y$ can be provided along with the camera's extrinsics ($\mathbf{R},\mathbf{T}$), to generate the target views corresponding to the specified poses. For the sake of brevity, we refer to this version of camera controllable LVDM as LVDM ($\mathcal{G}$) unless mentioned otherwise. 

\subsection{Semantic Uncertainty Quantification}
\label{subsec:uq}
\begin{figure}
    \centering
    \includegraphics[width=\linewidth]{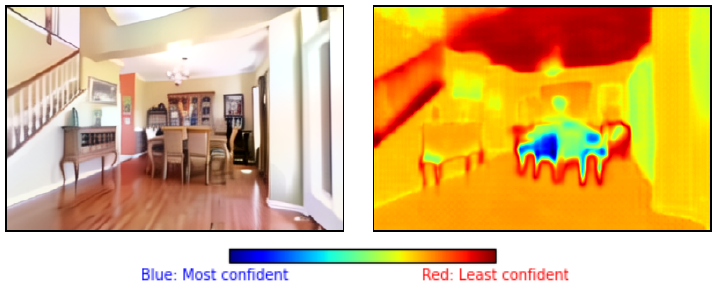}
    \caption{\textbf{Uncertainty Map Estimation.} On the right is the obtained uncertainty map from Pseudo Ground truth Image $\widetilde{\mathcal{I}}_{\text{p}}$ (on the left). It is generated by the LVDM after applying FST~\cite{yang2020fda}. These maps are crucial for guiding the Gaussian refinement process to progressively focus more on the confident ({\color[HTML]{0000FF}{blue}} and {\color[HTML]{009901}{green}} regions) and not on blurry ceiling and stairs ({\color[HTML]{ff0000}{red}}).}
    \vspace{-1em}
    \label{fig:uq}
\end{figure}

Even though the pseudo-images generated by the video diffusion model are plausible, they can generate over-saturated textures which adversely affects novel view synthesis (NVS) tasks (refer to {\hyperref[fig:ablation]{Figure~\ref{fig:ablation}}}). Since we are directly optimizing the Gaussian parameters representing the scene, the aforementioned issues will lead to the degradation of the composition of the scenes during refinement, resulting in unrealistic renderings. To this end, to make the LVDM self-aware of its predictions and pass only the confident pixel information during refinement, we utilize a semantic understanding-driven uncertainty quantification module as described below.\\ 

\noindent\textbf{Uncertainty Distillation.} 
For each pseudo-view $\widetilde{\mathcal{I}}_{\text{p}}$ generated by the LVDM, we use a Multimodal Large Language Model (MLLM)~\cite{li2023blip} to detect objects (even if partially visible) and obtain their object tags $O_{1\dots N}$. These $N{+}1$ classes (including background) are then passed to an open-vocabulary segmentation model $\mathcal{S}$~\cite{li2022languagedriven}, producing pixel-level logits:

\begin{equation}
o_t = \mathcal{S}\Bigl(\widetilde{\mathcal{I}}_{\text{p}},\, O\Bigr) \in \mathbb{R}^{n \times (N+1)},
\end{equation}

\noindent where $n$ is the number of pixels. 

Let \(p_{i,j,c}\) denote the predicted probability that the pixel at \((i,j)\) belongs to class \(c\). The per-pixel Shannon entropy is defined as:

\begin{equation}
\mathcal{U}_{(i,j)} = -\sum_{c} p_{i,j,c}\,\log\bigl(p_{i,j,c}\bigr),
\label{eq:entropy}
\end{equation}

\noindent which forms the uncertainty map \(\mathcal{U} \in \mathbb{R}^{H \times W}\) which is then subsequently used to perform pixel-wise multiplication with the LVDM image during refinement. Intuitively, $\mathcal{U}_{(i, j)}$ captures the uncertainty present in the segmentation map, whereby a higher entropy indicates a more uniform (i.e. less confident) class distribution for a pixel, while a lower entropy implies higher confidence in the predicted semantic classes. This entropy-based uncertainty measure thus provides a principled way to assess the reliability of the pseudo-view segmentation for our uncertainty-aware refinement. Further implementation details on how the MLLM was used for object tagging alongwith an open-set segmentation model~\cite{li2022languagedriven} is explained in the supplementary.\\

\noindent\textbf{Uncertainty-Weighted Reconstruction.} 
Instead of relying solely on the standard reconstruction loss introduced in {\hyperref[subsec:coarse_init]{Section~\ref{subsec:coarse_init}}}, we leverage the derived per-pixel uncertainty map $\mathcal{U}_{(i, j)}$ to guide the refinement process. In particular, given the rendered image $\mathcal{\tilde{I}}$ from the pre-trained reconstruction model $\mathcal{F}(\cdot)$ and the pseudo image $\widetilde{\mathcal{I}}_{\text{p}}$, we modify {\hyperref[eq:f3d_recon]{Equation~\ref{eq:f3d_recon}}} to compute an uncertainty-weighted reconstruction loss by taking the Hadamard (pixel-wise) product between the reconstruction error and the complement of the uncertainty map to get the pixel-wise confidence:
\begin{equation}
\resizebox{0.9\linewidth}{!}{$
\mathcal{L}_{\mathrm{uw\text{-}rec}} 
= \Bigl\| \bigl(1 - \mathcal{U}_{(i,j)}\bigr) \odot \bigl(\mathcal{\tilde{I}} - \widetilde{\mathcal{I}}_{\text{p}}\bigr) \Bigr\|_2 + \alpha\,\mathrm{SSIM}\bigl(\mathcal{\tilde{I}}, \widetilde{\mathcal{I}_{\text{p}}}\bigr)
$},
\label{eq:uw_recon}
\end{equation}

\noindent where $\odot$ denotes element-wise multiplication. This loss formulation ensures that regions with high segmentation uncertainty $U(i,j)$ contribute less to the overall loss, thus emphasizing the confident areas in the image. The optimization is steered towards refining the scene Gaussians in regions where the pseudo-view segmentation is reliable, leading to more accurate and visually coherent reconstruction.

\subsection{Iterative Refinement of Gaussians}
\label{subsec:refinement}

After obtaining the coarse Gaussians $\gamma_n$ from $\mathcal{F(\cdot)}$ in {\hyperref[subsec:coarse_init]{Section~\ref{subsec:coarse_init}}}, utilizing camera controlled LVDM $\mathcal{G}$ for generating pseudo views for guidance (in {\hyperref[subsec:lvdm_cam]{Section~\ref{subsec:lvdm_cam}}}) and defining our uncertainty weighted refinement objective in \hyperref[subsec:uq]{Section~\ref{subsec:uq}}, we optimize the parameters of $\gamma_n$ iteratively to obtain $\gamma^{'}_n$. However, it is observed that naive refinement of all the parameters leads to distorted renderings. We discuss the optimization strategy, style, texture alignment and the final refining objective for our pipeline below.

\noindent\textbf{Selective Optimization of Scales.} We observe that naive refinement of Gaussian scales obtained from the baseline method $\mathcal{F}(\cdot)$ can lead to an explosion in gradients and distorted renderings. This is primarily due to the tendency to fit multiple small Gaussians per pixel during the warm-up phase which rapidly increases the total number of Gaussians representing the scene. We limit this by optimizing the scales within a fixed range which we vary in an order of magnitude of $10^{-2}$ from the initial scales obtained from $\mathcal{F}(\cdot)$. 

\begin{figure*}[t!]
    \centering
    \includegraphics[width=0.9\linewidth]{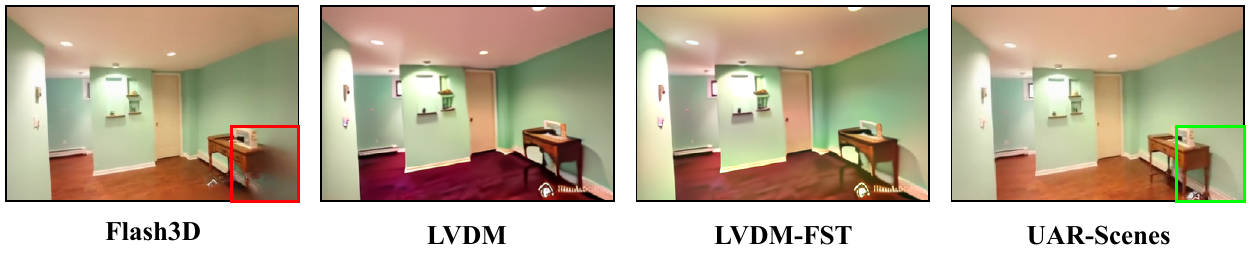}
    \vspace{-1em}
    \caption{\textbf{Ablation Results.} The leftmost image is the rendered view from the baseline method Flash3D which fails in extrapolation. Next, we have the LVDM generated image which clearly has oversaturated textures which does not align with real world scenes. On the 3rd image from the left, FST alleviates this issue by performing style alignment which leads to better quality results in the final output on the right.}
    \label{fig:ablation}
\end{figure*}

\noindent \textbf{Texture alignment with FST ($\Phi$).} LVDMs tend to exhibit over-saturated colors and features which may become more pronounced in indoor scenes (as shown in {\hyperref[fig:ablation]{Figure~\ref{fig:ablation}}}). While the generated images may be plausible, they might hurt the performance quantitatively in metrics like PSNR and SSIM which evaluate pixel-level accuracy. To address this, we perform \emph{Fourier Style Transfer (FST)} adaptation~\cite{yang2020fda} on the fly between the input image $\mathcal{I}$ and the output images generated by the LVDM model. This helps align the pseudo-images $\widetilde{\mathcal{I}}_{\text{p}}$ with the texture of the incoming source images. Aligning textures in this manner contributes to improving the overall performance, as shown in {\hyperref[tab:ablation]{Table~\ref{tab:ablation}}}.

\noindent\textbf{Refinement Objective.} During the iterative refinement stage, we adjust the parameters of each Gaussian primitive, \(\gamma_n = (\mu_n, R_n, S_n, \alpha_n)\), to better align the synthesized pseudo target view with the reference. Specifically, we optimize using the uncertainty-weighted reconstruction loss (Eq.\ref{eq:uw_recon}), \(\mathcal{L}_{\mathrm{uw\text{-}rec}}\), after performing the color correction with FST, which leads to our final optimization objective. 
\begin{equation}
\resizebox{0.9\linewidth}{!}{$
\mathcal{L}^{\Phi}_{\mathrm{uw\text{-}rec}} 
= \Bigl\| \bigl(1 - U(i,j)\bigr) \odot \bigl(\mathcal{\tilde{I}} - \Phi(\widetilde{\mathcal{I}}_{\text{p}}\bigr)) \Bigr\|_2 + \alpha\,\mathrm{SSIM}\bigl(\mathcal{\tilde{I}}, \Phi(\widetilde{\mathcal{I}}_{\text{p}})\bigr)
$},
\label{eq:uw_fst__recon}
\end{equation}
In practice, the Gaussian parameters are iteratively refined via a gradient-based scheme, ensuring that each update improves the fidelity of the rendered view while accounting for the uncertainty maps obtained using the segmentation model $\mathcal{S}$. The 2D supervisory updates rule progressively accounts for regions with higher confidence (lower uncertainty) that have a greater influence on the parameter updates, leading to refined, more accurate representations of the scene.

\section{Experiments}
\label{sec:expts}

We evaluate \name{} on the task of Novel View Synthesis (NVS) of scenes. Keeping in mind that multi-view consistency is a major focus to assess the quality of refined scenes, we present results on both narrow and wide baselines in {\hyperref[tab:re10k_nvs]{Table~\ref{tab:re10k_nvs}}}. We further evaluate the interpolation and extrapolation performance following~\cite{szymanowicz2024flash3d} and \cite{wewer2024latentsplat} after refinement in {\hyperref[tab:interp_extrap]{Table~\ref{tab:interp_extrap}}}. To highlight the transferability of our method across datasets, we also show NVS results on KITTI-v2 in {\hyperref[tab:kitti]{Table~\ref{tab:kitti}}}. This shows our refinement pipeline can be seamlessly integrated with any existing reconstruction method across a wide variety of tasks. Finally, we present ablation studies in showing the importance of each component in our framework and how it contributes towards overall performance.

\begin{figure*}
    \centering
    \includegraphics[width=0.9\textwidth]{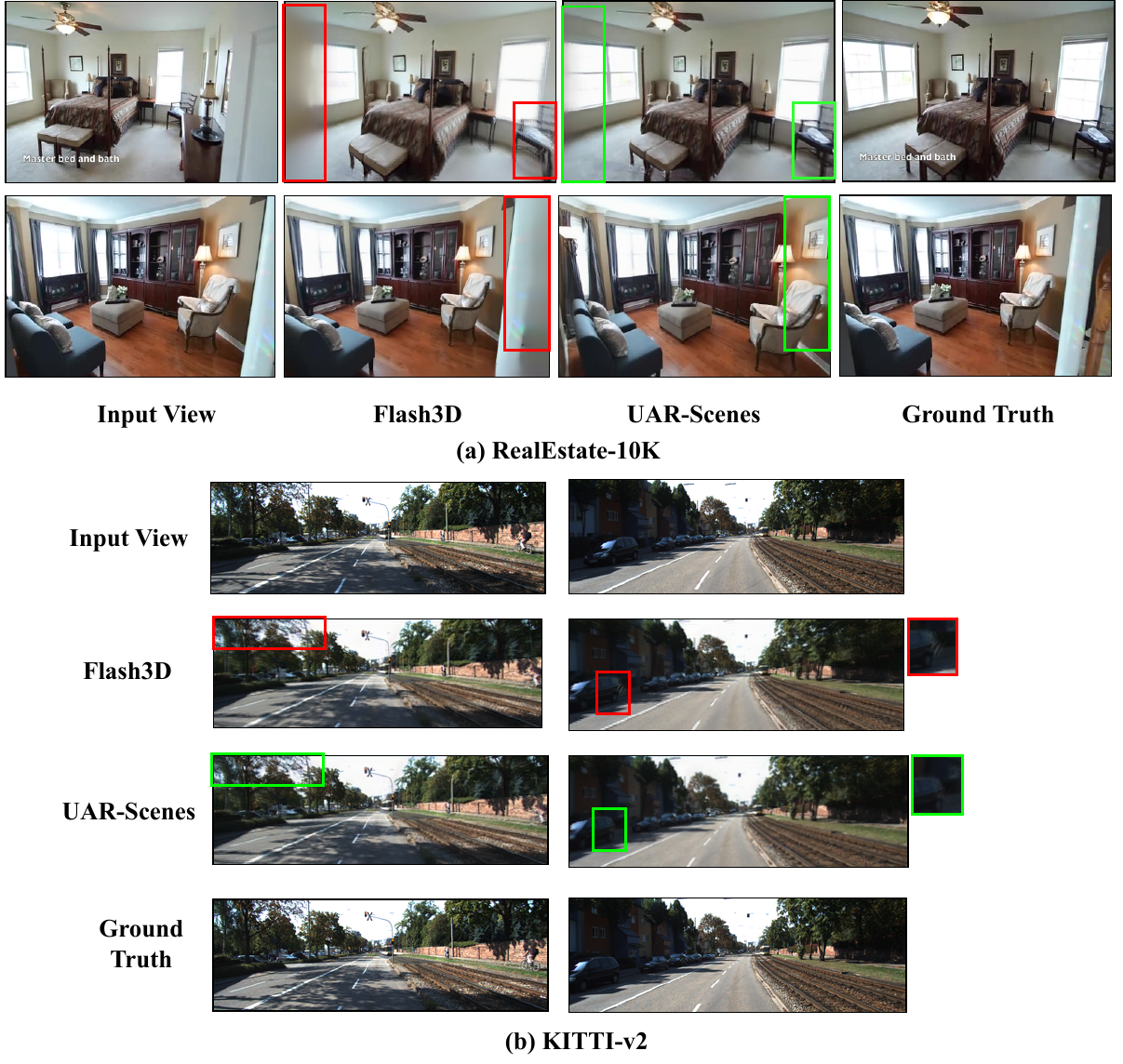}
    \vspace{-1em}
    \caption{\textbf{Qualitative Results.} (a) Qualitative comparisons on the RealEstate-10K dataset shows that our method produces more \emph{realistic results} which are more plausible and faithful to the original image (In 1st row, Flash3D's renderings are blurry outside the camera's seen frustum where \name{} is able to complete the window. Similarly \name{} can provide reasonable completions which may not always align with the GT as shown in 2nd row). (b) Qualitative comparison on the KITTI-v2 dataset which shows that our method can deliver \emph{sharp results especially in edges} where there may be ambiguity (Back edge of car is distorted in Flash3D's prediction in 2nd column). Notice that despite the significant camera motion between the original input view and the target novel views, \name{} can render realistic and plausible renderings as highlighted above.}
    \vspace{-1em}
    \label{fig:qualitative}
\end{figure*}

\begin{table*}[t]
\centering
\caption{\textbf{Novel View Synthesis.} Our model shows superior performance on RealEstate10k~\cite{zhou2018stereo} for small, medium, and large baseline ranges. We highlight the best performance in \textbf{bold} and the second best performance in \underline{underline}.}
\label{tab:re10k_nvs}
\resizebox{0.8\textwidth}{!}{%
\begin{tabular}{l ccc ccc ccc}
\toprule
\multirow{2}{*}{Model} & \multicolumn{3}{c}{5 frames} & \multicolumn{3}{c}{10 frames} & \multicolumn{3}{c}{$\mathcal{U}$[-30,30]} \\
\cmidrule(lr){2-4}\cmidrule(lr){5-7}\cmidrule(lr){8-10}
 & PSNR$\uparrow$ & SSIM$\uparrow$ & LPIPS$\downarrow$
 & PSNR$\uparrow$ & SSIM$\uparrow$ & LPIPS$\downarrow$
 & PSNR$\uparrow$ & SSIM$\uparrow$ & LPIPS$\downarrow$ \\
\midrule
Syn-Sin~\cite{wiles2020synsin}         & --    & --    & --    & --    & --    & --    & 22.30 & 0.740 & -- \\
SV-MPI~\cite{tucker2020single}          & 27.10 & 0.870 & --    & 24.40 & 0.812 & --    & 23.52 & 0.785 & -- \\
BTS~\cite{wimbauer2023behind}            & --    & --    & --    & --    & --    & --    & 24.00 & 0.755 & 0.194 \\
Splatter Image~\cite{szymanowicz2024splatter}  & 24.15 & 0.894 & 0.110 & 25.60 & 0.760 & 0.240 & 23.10 & 0.730 & 0.290 \\
MINE~\cite{li2021mine}                  & 28.45 & 0.897 & 0.111 & 25.89 & 0.850 & 0.150 & 24.75 & 0.820 & 0.179 \\
Flash3D~\cite{szymanowicz2024flash3d}   & \underline{28.46} & \underline{0.899} & \underline{0.100} & \underline{25.94} & \underline{0.857} & \underline{0.133} & \underline{24.93} & \underline{0.833} & \underline{0.160} \\
\midrule
\name{} & \textbf{28.67} & \textbf{0.902} & \textbf{0.095} 
                & \textbf{26.54} & \textbf{0.861} & \textbf{0.112} 
                & \textbf{27.81} & \textbf{0.887} & \textbf{0.107} \\
\bottomrule
\end{tabular}%
}
\end{table*}
\begin{table*}[t]
\centering
\vspace{-0.5em}
\caption{\textbf{Interpolation vs.\ Extrapolation.} We compare our method (\name{}) on the RealEstate-10K dataset against baselines on PSNR, SSIM, LPIPS, and FID metrics. We highlight the best performance in \textbf{bold} and the second best performance in \underline{underline}.}
\small 
\setlength{\tabcolsep}{4pt} 
\resizebox{0.675\textwidth}{!}{%
\begin{tabular}{l ccc cccc}
\toprule
\multirow{2}{*}{Method} & \multicolumn{3}{c}{Interpolation} & \multicolumn{4}{c}{Extrapolation} \\
\cmidrule(lr){2-4}\cmidrule(lr){5-8}
 & PSNR$\uparrow$ & SSIM$\uparrow$ & LPIPS$\downarrow$
 & PSNR$\uparrow$ & SSIM$\uparrow$ & LPIPS$\downarrow$ & FID$\downarrow$ \\
\midrule
PixelNeRF~\cite{yu2021pixelnerf}      & 24.00 & 0.589 & 0.550 & 20.05 & 0.575 & 0.567 & 160.77 \\
Du et al.~\cite{du2023learning}         & 24.78 & 0.820 & 0.410 & 21.23 & 0.760 & 0.480 & 14.34  \\
pixelSplat~\cite{charatan2024pixelsplat} & 25.49 & 0.794 & 0.291 & 22.62 & 0.777 & 0.216 & 5.78   \\
latentSplat~\cite{wewer2024latentsplat}   & 25.53 & 0.853 & 0.280 & 23.45 & 0.801 & 0.190 & \underline{2.97}   \\
MVSplat~\cite{chen2024mvsplat}            & \textbf{26.39} & \underline{0.869} & \underline{0.128} & 24.04 & 0.812 & 0.185 & 3.87   \\
Flash3D~\cite{szymanowicz2024flash3d}     & 23.87 & 0.811 & 0.185 & \underline{24.10} & \underline{0.815} & \underline{0.185} & 4.02   \\
\midrule
\name{}                        & \underline{26.37} & \textbf{0.871} & \textbf{0.125} & \textbf{24.37} & \textbf{0.819} & \textbf{0.144} & \textbf{2.55} \\
\bottomrule
\end{tabular}%
}
\label{tab:interp_extrap}
\end{table*}

\begin{table}[t]
\centering
\vspace{-1em}
\caption{\textbf{Out-Domain Evaluation.} We evaluate on the KITTI-v2~\cite{geiger2012we} Dataset. We highlight the best performance in \textbf{bold} and the second best performance is \underline{underlined}. We beat the baseline method comprehensively.}
\label{tab:kitti}
\resizebox{0.75\columnwidth}{!}{%
\begin{tabular}{l c c c}
\toprule
& \multicolumn{3}{c}{\textbf{KITTI}} \\
\cmidrule(lr){2-4}
\textbf{Method} & \textbf{PSNR}$\uparrow$ & \textbf{SSIM}$\uparrow$ & \textbf{LPIPS}$\downarrow$ \\
\midrule
LDI~\cite{tulsiani2018layer}    & 16.50 & 0.572 & -- \\
SV-MPI~\cite{tucker2020single}  & 19.50 & 0.733 & -- \\
BTS~\cite{wimbauer2023behind}   & 20.10 & 0.761 & 0.144 \\
MINE~\cite{li2021mine}          & 21.90 & 0.828 & 0.112 \\
Flash3D~\cite{szymanowicz2024flash3d} & \underline{21.96} & \underline{0.826} & \underline{0.132} \\
\midrule
\name{}               & \textbf{22.31} & \textbf{0.844} & \textbf{0.128} \\
\bottomrule
\end{tabular}%
}
\vspace{-1em}
\end{table}

\subsection{Experimental Setup}
\textbf{Datasets.} Our experiments involve in-domain results on the real-world scenes dataset RealEstate-10K~\cite{zhou2018stereo} and out-domain results on the driving dataset KITTI-v2~\cite{geiger2012we} following the protocol of the base reconstruction model~\cite{szymanowicz2024flash3d}. RealEstate-10K consists of 69893 real estate videos showing the indoor and outdoor views of houses collected from YouTube. KITTI-V2 consists of driving sequences recorded with a stereo camera per day. We follow the standard protocol of 1079 images for testing this dataset.

\noindent\textbf{Evaluation Metrics.} We adopt PSNR, SSIM~\cite{wang2004image} and Perceptual Similarity (LPIPS)~\cite{zhang2018unreasonable} as our photo-metric evaluation standard as in~\cite{charatan2024pixelsplat, chen2024mvsplat, szymanowicz2024flash3d}. Further, we report Frechet-Inception Distance (FID)~\cite{heusel2017gans} as well since we generate plausible explanations for unobserved regions which do not have a corresponding paired ground truth.

\noindent\textbf{Pseudo-View Generation.} We utilize the MotionCtrl~\cite{wang2024motionctrl} LVDM model which has fine-grained camera control for generating the pseudo views and provides supervision for refining the Gaussian primitives. This is essential to align the generated views with those obtained by rendering the coarse Gaussians of $\mathcal{F(\cdot)}$. Following this, we perform 50 inference sampling steps with a reduced batch size of 10 to address the increased computational load. 

\noindent\textbf{Optimization Details.} We optimize each scene for 1000 steps in which the learning rate of position information decays from $1 \times 10^{-3}$ to $2 \times 10^{-5}$.  We use a single NVIDIA L40 48 GB GPU for running all our experiments. Finally, our rendering resolution for all images is $256 \times 384$ following the protocol in~\cite{szymanowicz2024flash3d}. We keep a batch size of 2 for all our experiments. Further implementation details are listed in the supplementary material.

\subsection{Novel View Synthesis} 
\textbf{RealEstate-10K.} We benchmark our refinement model on the RealEstate dataset where we show improvements over stereo and sparse view methods while refining a single-image to 3D reconstruction pipeline. \name{} obtains around 1dB improvement on average across both the closer and wide baselines in PSNR showing the effectiveness of our pipeline in improving existing feed-forward models. Notice that while the performance of the baseline model Flash3D steadily decreases as the distance from the source increases, \name{} is able to still provide decent renderings. This is important as it suggests that these feed-forward methods fail to capture those areas of the scene which progressively start falling out of the range of the input conditioning image. We provide qualitative results in {\hyperref[fig:teaser]{Figure~\ref{fig:teaser}}} and {\hyperref[fig:qualitative]{Figure~\ref{fig:qualitative}}(a)} where we clearly show how \name{} provides high quality rendering results in cases where Flash3D fails. We show additional qualitative results in the supplementary.\\
\textbf{KITTI-v2.} We also perform an out-domain evaluation on the KITTI dataset showing that our refinement procedure is dataset agnostic and can be used in a post-hoc fashion to adapt to any setting. We obtain a PSNR improvement of around 0.35 dB over the baseline method. We provide further qualitative results in {\hyperref[fig:qualitative]{Figure~\ref{fig:qualitative}}(b)}.

\subsection{Generative Results} We report interpolation and extrapolation results on the RealEstate-10K dataset using Flash3D as the baseline method. We are able to beat all the methods except MVSplat which uses stereo information to interpolate. In Extrapolation, over a wide baseline, our FID is significantly lower then all methods including LatentSplat~\cite{wewer2024latentsplat} which uses a generative GAN denoiser but still fails to handle complex real-world scenes. Except Flash3D~\cite{szymanowicz2024flash3d}, we report FID numbers following existing feed forward methods~\cite{wewer2024latentsplat, chen2024mvsplat360}.

\subsection{Ablation Studies}
\label{subsec:ablation}
\begin{table}[t]
\centering
\vspace{-1em}
\caption{\textbf{Ablation Studies.} We report PSNR ($\uparrow$), SSIM ($\uparrow$), and LPIPS ($\downarrow$) on RealEstate-10K. The component columns indicate whether LVDM ($\mathcal{G}$), FST ($\Phi$), and uncertainty ($\mathcal{U}$) are included.}
\label{tab:ablation}
\small
\setlength{\tabcolsep}{2pt} 
\begin{tabular}{l c c c c c c}
\toprule
\multirow{2}{*}{\textbf{Models}} & \multicolumn{3}{c}{\textbf{Components}} & \multicolumn{3}{c}{\textbf{Metrics}} \\
\cmidrule(lr){2-4}\cmidrule(lr){5-7}
 & $\mathcal{G}$ & $\Phi$ & $\mathcal{U}$ & PSNR$\uparrow$ & SSIM$\uparrow$ & LPIPS$\downarrow$ \\
\midrule
Baseline                    & \xmark & \xmark & \xmark & 24.93 & 0.833 & 0.160 \\
Baseline + LVDM              & \cmark & \xmark & \xmark & 27.24 & 0.867 & 0.126 \\
Baseline + LVDM-FST        & \cmark & \cmark & \xmark & 27.33 & 0.869 & 0.119 \\
\name{} & \cmark & \cmark & \cmark & \textbf{27.81} & \textbf{0.887} & \textbf{0.107} \\
\bottomrule
\end{tabular}
\vspace{-2em}
\end{table}
\textbf{Architectural Choice.} We conduct our ablation studies on the real world RealEstate-10K dataset to measure the effects of the various architectural designs on the overall performance. The results are listed in {\hyperref[tab:ablation]{Table~\ref{tab:ablation}}}. It can be observed that adding the LVDM and performing uncertainity aware refinement contributes significantly to performance gains. The rendering effect behind each choice is shown in {\hyperref[fig:ablation]{Figure~\ref{fig:ablation}}}. Note how the oversaturated texture information generated by the LVDM does not align with either the input image or the ground truth. We therefore perform the alignment operation using $\Phi$ to produce better quality supervision which is more practical for Novel View Synthesis tasks for scenes.
\vspace{-0.5em}
\section{Conclusion}
\label{sec:conclusion}
We introduced \name{}, a novel 3D scene refinement pipeline that enhances scene Gaussians derived from a single image, thereby improving the quality of Gaussians produced by 3D reconstruction pipelines. Our results demonstrate the versatility of \name{} across both in-domain and out-domain datasets. Through Novel View Synthesis experiments, we outperform state-of-the-art feed-forward methods across small, medium, and large baseline settings. Additionally, \name{} excels in challenging interpolation and extrapolation tasks, yielding superior rendered views and high-quality generations at unseen camera angles, as evidenced by both qualitative and quantitative results. Furthermore, our ablation studies validate the effectiveness of individual components and design choices within the \name{} pipeline, highlighting the benefits of employing Fourier-style texture alignment with real-world scenes. Overall, our findings highlight \name{}'s potential to advance 3D scene understanding and novel view synthesis.\\
\noindent
\textbf{Acknowledgments:} We gratefully acknowledge the support of USDA NRI grant 2021-67022-33453, UC MRPI grant A21-0101-S003, and NSF grant CMMI-2326309. Any opinions, findings, and conclusions or recommendations expressed in this material are those of the authors and do not necessarily reflect the views of the funding agencies.

{\small
\bibliographystyle{unsrt}
\bibliography{09_references}
}
\end{document}